\documentclass[10pt,twocolumn,letterpaper]{article}
\usepackage{multirow}
\usepackage{cvpr}
\usepackage{times}
\usepackage{epsfig}
\usepackage{graphicx}
\usepackage{amsmath}
\usepackage{amssymb}
\usepackage{amsthm}
\usepackage{tabularx}
\usepackage{color}

\usepackage{dsfont}
\usepackage{multirow}
\usepackage{booktabs}
\usepackage{algorithm, a lgorithmic}
\usepackage{rotating}
\usepackage{pifont}
\newcommand{\cmark}{\ding{51}}%

\newcommand{\dd}{\mathop{}\,\mathrm{d}}
\newcommand{\cn}{\mathcal{N}}

\usepackage[pagebackref=true,breaklinks=true,letterpaper=true,colorlinks,bookmarks=false]{hyperref}

\cvprfinalcopy 

\def\cvprPaperID{****} 

\ifcvprfinal\fi
\begin{document}

\title{A Variational Auto-Encoder Model for Stochastic Point Processes}

\author{Nazanin Mehrasa$^{1,3}$, Akash Abdu Jyothi$^{1,3}$, Thibaut Durand$^{1,3}$, Jiawei He$^{1,3}$,  Leonid Sigal$^{2,3}$, Greg Mori$^{1,3}$ \\
$^{1}$Simon Fraser University \qquad $^2$University of British Columbia \qquad $^{3}$Borealis AI\\
{\tt\small \{nmehrasa, aabdujyo, tdurand, jha203\}@sfu.ca}\qquad
{\tt\small{lsigal}@cs.ubc.ca}\qquad
{\tt\small{mori}@cs.sfu.ca}
}
\maketitle
\thispagestyle{empty}

\begin{abstract}
We propose a novel probabilistic generative model for action sequences.  The model is termed the Action Point Process VAE (APP-VAE), a variational auto-encoder that can capture the distribution over the times and categories of action sequences.  Modeling the variety of possible action sequences is a challenge, which we show can be addressed via the APP-VAE's use of latent representations and non-linear functions to parameterize distributions over which event is likely to occur next in a sequence and at what time.  We empirically validate the efficacy of APP-VAE for modeling action sequences on the MultiTHUMOS and Breakfast datasets.
\end{abstract}


\section{Introduction}
\label{sec:introduction}

\begin{figure}[t]
\includegraphics[width=0.97\linewidth]{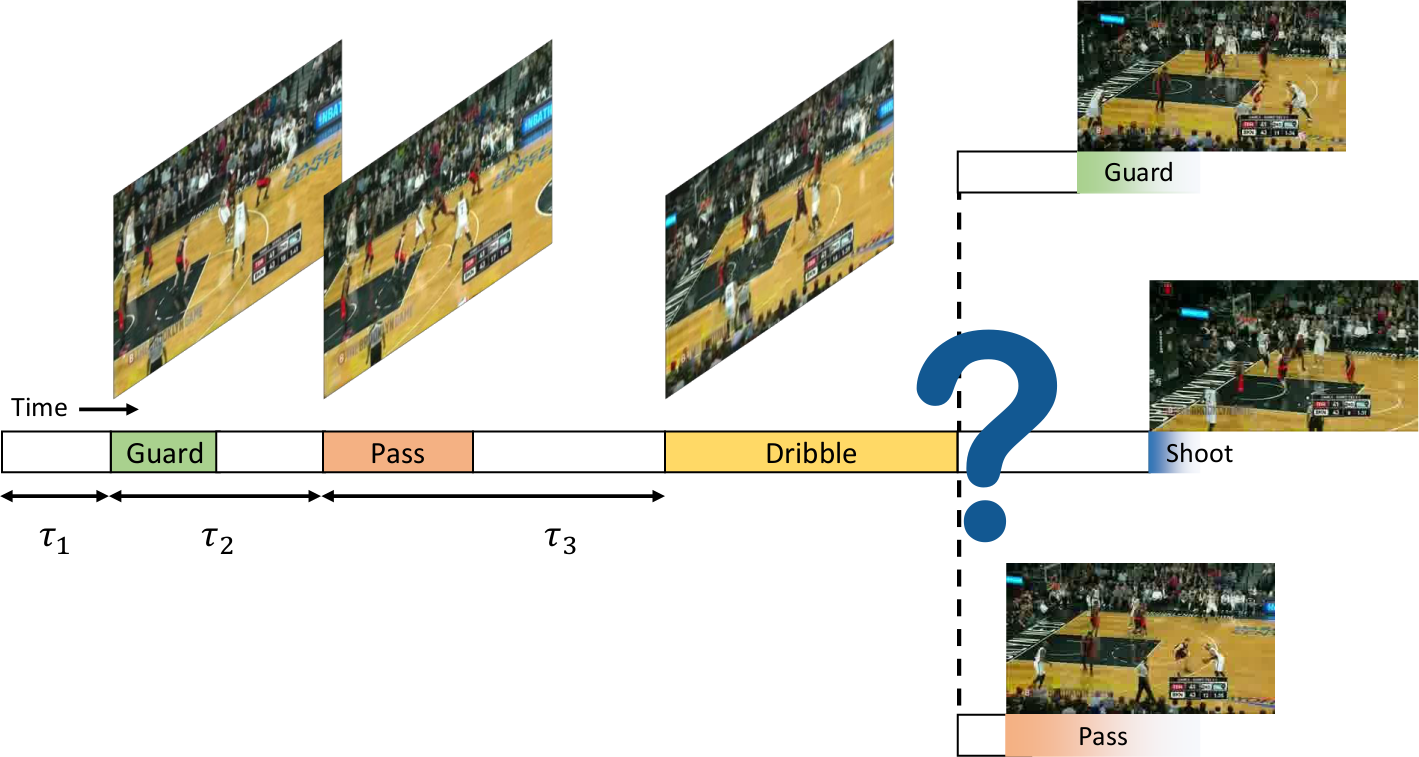} \\
\caption{It is difficult to make predictions, especially about the future.  Given a history of past actions, multiple actions are possible in the future. We focus on the problem of learning a distribution over the future actions -- \textit{what} are the possible action categories and \textit{when} will they start. }
\label{fig:pull}
\end{figure}

Anticipatory reasoning to model the evolution of action sequences over time is a fundamental challenge in human activity understanding.  The crux of the problem in making predictions about the future is the fact that for interesting domains, the future is uncertain -- given a history of actions such as those depicted in Fig.~\ref{fig:pull}, the distribution over future actions has substantial entropy.

In this work, we propose a powerful generative approach that can effectively model the categorical and temporal variability comprising action sequences.
Much of the work in this domain has focused on taking frame level data of video as input in order to predict the actions or activities that may occur in the immediate future. There has also been recent interest on the task of predicting the sequence of actions that occur farther into the future \cite{kdd_action_prediction,2018arXiv180804063Z,Farha2018}.

Time series data often involves regularly spaced data points with interesting events occurring sparsely across time. This is true in case of videos where we have a regular frame rate but events of interest are present only in some frames that are infrequent. 
We hypothesize that in order to model future events in such a scenario, it is beneficial to consider the history of sparse events (action categories and their temporal occurrence in the above example) alone, instead of regularly spaced frame data. While the history of frames contains rich information over and above the sparse event history, we can possibly create a model for future events occurring farther into the future by choosing to only model the sparse sequence of events. 
This approach also allows us to model high-level semantic meaning in the time series data that can be difficult to discern from low-level data points that are regular across time. 

Our model is formulated in the variational auto-encoder (VAE)~\cite{Kingma2014} paradigm, a powerful class of probabilistic models that facilitate generation and the ability to model complex distributions.  We present a novel form of VAE for action sequences under a point process approach.  This approach has a number of advantages, including a probabilistic treatment of action sequences to allow for likelihood evaluation, generation, and anomaly detection.

\begin{figure*}[t]
\centering
\includegraphics[scale=.4]{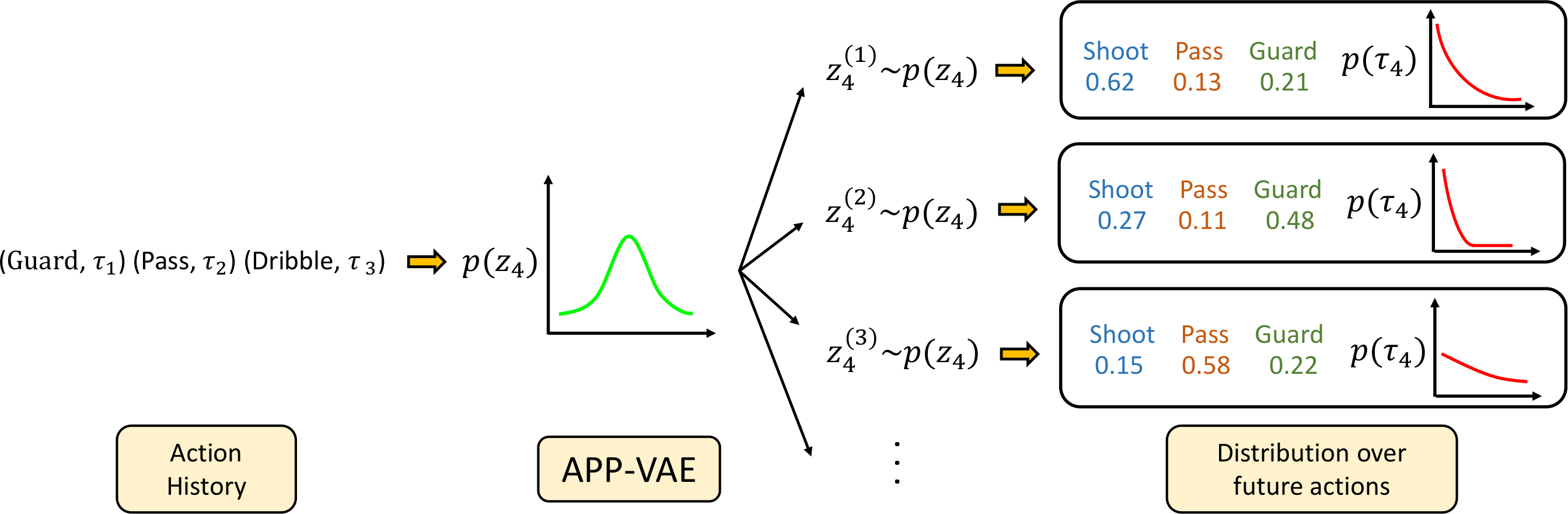}
\caption{Given the history of actions, APP-VAE generates a distribution over possible actions in the next step. APP-VAE can recurrently perform this operation to model diverse sequences of actions that may follow. The figure shows the distributions for the fourth action in a basketball game given the history of first three actions.}
\label{fig:fig2}
\end{figure*}

\textbf{Contribution.} The contributions of this work center around the APP-VAE (Action Point Process VAE), a novel generative model for asynchronous time action sequences.  The contributions of this paper include:
\begin{itemize}
    \item A novel formulation for modeling point process data within the variational auto-encoder paradigm.
    \item Conditional prior models for encoding asynchronous time data.
    \item A probabilistic model for jointly capturing uncertainty in which actions will occur and when they will happen.
\end{itemize}

\section{Related Work}
\label{sec:related_works} 

\textbf{Activity Prediction.}
Most activity prediction tasks are frame-based, \ie the input to the model is a sequence of frames before the action starts and the task is predict what will happen next.  
Lan \etal~\cite{10.1007/978-3-319-10578-9_45} predict future actions from hierarchical representations of short clips by having different classifiers at each level in a max-margin framework. 
Mahmud \etal~\cite{Mahmud_2017_ICCV} jointly predicts future activity as well as its starting time by a multi-streams framework. Each streams tries to catch different features for having a richer feature representation for future prediction: One stream for visual information, one for previous activities and the last one focusing on the last activity.

Farha \etal~\cite{Farha2018} proposed a framework for predicting the action categories of a sequence of future activities as well as their starting and ending time. They proposed two deterministic models, one using a combination of RNN and HMM and the other one is a CNN predicting a matrix which future actions are encoded in it. 

\textbf{Asynchronous Action Prediction.}  We focus on the task of predicting future action given a sequence of previous actions that are asynchronous in time.
Du \etal~\cite{kdd_action_prediction} proposed a recurrent temporal model for learning the next activity timing and category given the history of previous actions. Their recurrent model learns a non-linear map of history to the intensity function of a temporal point process framework. Zhong \etal~\cite{2018arXiv180804063Z} also introduced a hierarchical recurrent network model for future action prediction for modeling future action timing and category. Their model takes frame-level information as well as sparse high-level events information in the history to learn the intensity function of a temporal point process. Xiao \etal~\cite{xiao2017wasserstein} introduced an intensity-free generative method for temporal point process. The generative part of their model is an extension of Wasserstein GAN in the context of temporal point process for learning to generate sequences of action.

\textbf{Early Stage Action Prediction.}
Our work is related to early stage action prediction. This task refers to predicting the action given the initial frames of the activity~\cite{7780583,6248012,Shi_2018_ECCV}. Our task is different from early action prediction, because the model doesn't have any information about the action while predicting it. Recently Yu \etal~\cite{DBLP:journals/corr/abs-1711-09265} used variational auto-encoder to learn from the frames in the history and transfer them into the future. Sadegh Aliakbarian \etal~\cite{Aliakbarian2017} combine context and action information using a multi-stage LSTM model to predict future action. The model is trained with a loss function which encourages the model to predict action with few observations. Gao \etal~\cite{DBLP:journals/corr/GaoYN17aa} proposed to use a Reinforced Encoder-Decoder network for future activity prediction.  Damen \etal~\cite{butepage2018classify} proposed a semi-supervised variational recurrent neural network to model human activity including classification, prediction, detection and anticipation of human activities.

\textbf{Video Prediction.}
Video prediction has recently been studied in several works. Denton and Fergus~\cite{Denton2018} use a variational auto-encoder framework with a learned prior to generate future video frames. He \etal~\cite{He_2018_ECCV} also proposed a generative model for future prediction. They structure the latent space by adding control features which makes the model able to control generation. Vondrick \etal~\cite{Vondrick_2017_CVPR} uses adversarial learning for generating videos of future with transforming the past pixels.  Patraucean et al.~\cite{PatrauceanHC16} describe a spatio-temporal auto-encoder that predicts optical flow as a dense map, using reconstruction in its learning criterion.  Villegas et al.~\cite{villegas17hierchvid} propose a hierarchical approach to pixel-level video generation, reasoning over body pose before rendering into a predicted future frame.

\section{Asynchronous Action Sequence Modeling}
\label{sec:model}

\begin{figure*}[th]
\centering
\begin{tabular}{cc}

\rotatebox[origin=l]{90}{\textsc{~~~~~~~~Training}}&\includegraphics[scale=0.5]{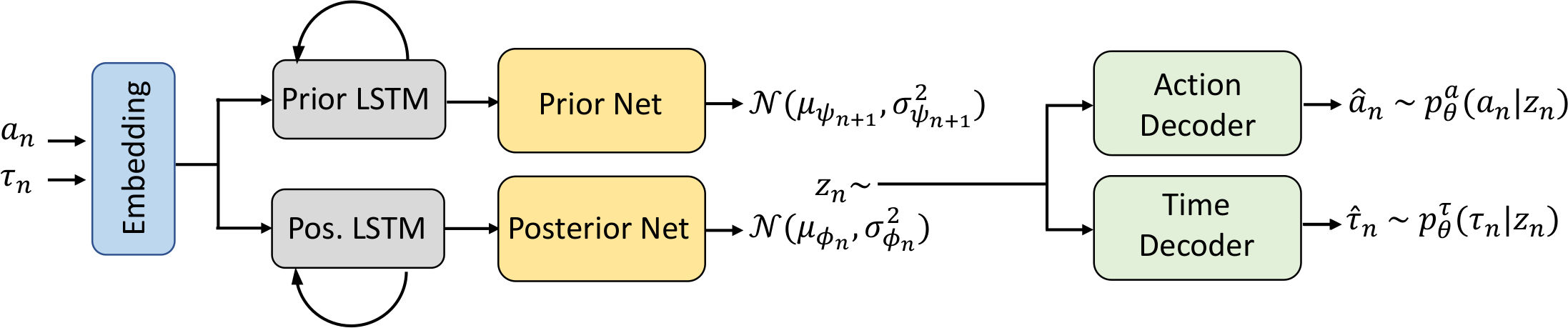} \\ 
\midrule
\rotatebox[origin=l]{90}{\textsc{~~Generation}}& \includegraphics[scale=0.5]{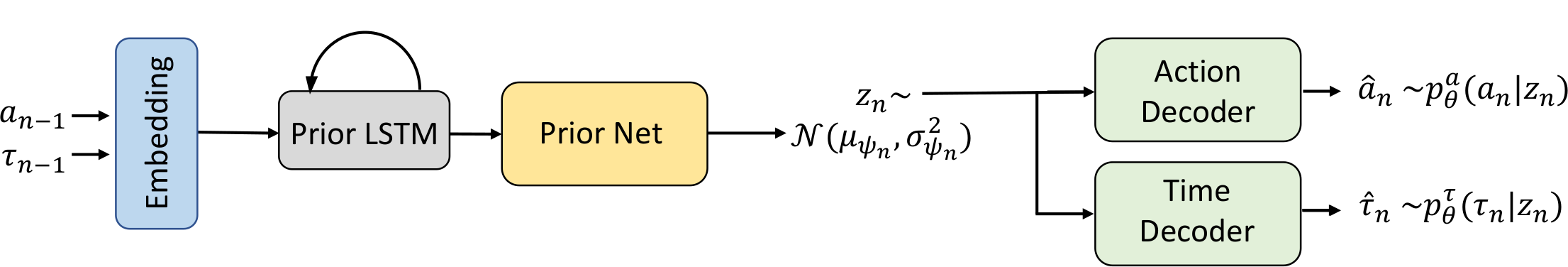} \\
\end{tabular}
\caption{Our proposed recurrent VAE model for asynchronous action sequence modeling. At each time step, the model uses the history of actions and inter-arrival times to generate a distribution over latent codes, a sample of which is then decoded into two probability distributions for the next action: one over possible action labels and one over the inter arrival time.}
\label{fig:arch}
\end{figure*}

We first introduce some notations and the problem definition.
Then we review the VAE model and temporal point process that are used in our model.  Subsequently, we present our model in detail and how it is trained.

\paragraph{Problem definition.}
The input is a sequence of actions $x_{1:n}=(x_1, \ldots, x_{n})$ where $x_n$ is the $n$-th action. 
The action $x_n=(a_n, \tau_n)$ is represented by the action category $a_n \in \{1, 2, \ldots, K\}$ ($K$ discrete action classes) and the inter-arrival time $\tau_n \in \mathbb{R}^{+}$. 
The inter-arrival time is the difference between the starting time of action $x_{n-1}$ and $x_n$.
We formulate the asynchronous action distribution modeling task as follows: given a sequence of actions $x_{1:n-1}$, the goal is to produce a distribution over what action $a_{n}$ will happen next, and the inter arrival time $\tau_n$. 
We aim to develop probabilistic models to capture the uncertainty over these what and when questions of action sequence modeling.

\subsection{Background: Base Models}
\paragraph{Variational Auto-Encoders (VAEs).}
A VAE \cite{Kingma2014} describes a generative process with simple prior $p_\theta(z)$ (usually chosen to be a multivariate Gaussian) and complex likelihood $p_{\theta}(x|z)$ (the parameters of which are produced by neural networks). $x$ and  $z$ are observed and latent variables, respectively. Approximating the intractable posterior $p_{\theta}(z|x)$ with a recognition neural network $q_{\phi}(z|x)$, the parameters of the generative model ${\theta}$ as well as the recognition model ${\phi}$ can be jointly optimized by maximizing the evidence lower bound $\mathcal{L}$ on the marginal likelihood $p_\theta(x)$:
\begin{align}
\begin{split}
\log p_\theta(x) &= \textrm{KL}(q_{\phi} \Vert p_{\theta}) + \mathcal{L}({\theta},{\phi})\\
&\geq \mathcal{L}({\theta},{\phi}) =-\mathbb{E}_{q_\phi}\left[\log \frac{q_\phi(z|x)}{p_\theta(z,x)}\right].
\end{split}
\end{align}

Recent works expand VAEs to time-series data including video \cite{Babaeizadeh2018,Denton2018,He_2018_ECCV}, text~\cite{chung2015recurrent,hu2017toward}, or audio~\cite{yingzhen2018disentangled}. A popular design choice of such models is the integration of a per time-step VAE with RNN/LSTM temporal modelling. The ELBO thus becomes a summation of time-step-wise variational lower bound\footnote{Note that variants exist, depending on the exact form of the recurrent structure and its VAE instantiation.}:
\begin{eqnarray}
    \mathcal{L}(\theta,\phi,\psi) &=& \displaystyle\sum_{n=1}^{N} \Big[ \mathbb{E}_{q_{\phi}(z_{1:n}|x_{1:n})}\left[\log  p_{\theta}(x_{n}|x_{1:n-1},z_{1:n})\right] \nonumber \\
    &-& \textrm{KL}(q_{\phi}(z_{n}|x_{1:n}) || p_{\psi}(z_{n} | x_{1:n-1})) \Big].
\label{eq:obj_func}
\end{eqnarray}
with a ``prior" $p_{\psi}(z_n | x_{1:n-1})$ that evolves over the $N$ time steps used.

\paragraph{Temporal point process.}
A temporal point process is a stochastic model used to capture the inter-arrival times of a series of events.
A temporal point process is characterized by the conditional intensity function $\lambda(\tau_n|x_{1:n-1})$, which is conditioned on the past events $x_{1:n-1}$ (\eg action in this work).
The conditional intensity encodes instantaneous probabilities at time $\tau$. 
Given the history of $n-1$ past actions, the probability density function for the time of the next action is:
\begin{align}
f(\tau_n|x_{1:n-1}) = \lambda(\tau_n|x_{1:n-1}) e ^ {- \int\limits_0^{\tau_n} \lambda(u|x_{1:n-1}) \dd u }
\end{align}
The Poisson process \cite{Kingman1993} is a popular temporal point process, which assumes that events occur independent of one another.
The conditional intensity is $\lambda(\tau_n|x_{1:n-1})=\lambda$ where $\lambda$ is a positive constant.
More complex conditional intensities have been proposed like Hawkes Process \cite{Hawkes1971} and Self-Correcting Process \cite{Isham1979}.
All these conditional intensity function seek to capture some forms of dependency on the past action.   
However, in practice the true model of the dependencies is never known \cite{Mei2017} and the performance depend on the design of the conditional intensity.
In this work, we learn a recurrent model that estimates the conditional intensity based on the history of actions.

\subsection{Proposed Approach}

We propose a generative model for asynchronous action sequence modeling using the VAE framework. 
\autoref{fig:arch} shows the architecture of our model. 
Overall, the input sequence of actions and inter arrival times are encoded using a recurrent VAE model.
At each step, the model uses the history of actions to produce a distribution over latent codes $z_n$, a sample of which is then decoded into two probability distributions: one over the possible action categories and another over the inter-arrival time for the next action.
We now detail our model.

\paragraph{Model.}
At time step $n$ during training, the model takes as input the action $x_n$, which is the target of the prediction model, and the history of past actions $x_{1:n-1}$.
These inputs are used to compute a conditional distribution $q_\phi(z_n|x_{1:n})$ from which a latent code $z_n$ is sampled.
Since the true distribution over latent variables $z_n$ is intractable we rely on a time-dependent inference network $q_\phi(z_{n}|x_{1:n})$ that approximates it with a conditional Gaussian distribution $\cn(\mu_{\phi_n}, \sigma^2_{\phi_n})$.
To prevent $z_n$ from just copying $x_n$, we force $q_\phi(z_n|x_{1:n})$ to be close to the prior distribution $p(z_n)$ using a KL-divergence term.
Usually in VAE models, $p(z_n)$ is a fixed Gaussian $\cn(0,I)$. 
But a drawback of using a fixed prior is that samples at each time step are drawn randomly, and thus ignore temporal dependencies present between actions.
To overcome this problem, a solution is to learn a prior that varies across time, being a function of all past actions except the current action $p_\psi(z_{n+1}|x_{1:n})$.
Both prior and approximate posterior are modelled as multivariate Gaussian distributions with diagonal covariance with parameters as shown below: 
\begin{align}
q_\phi(z_n|x_{1:n}) &= \mathcal{N}(\mu_{\phi_n}, \sigma^2_{\phi_n})
\\
p_\psi(z_{n+1}|x_{1:n}) &= \mathcal{N}(\mu_{\psi_{n+1}},\sigma^2_{\psi_{n+1}})
\end{align}
At step $n$, both posterior and prior networks observe actions $x_{1:n}$ but the posterior network outputs the parameters of a conditional Gaussian distribution for the current action $x_n$ whereas the prior network outputs the parameters of a conditional Gaussian distribution for the next action $x_{n+1}$.

At each time-step during training, a latent variable $z_n$ is drawn from the posterior distribution $q_\phi(z_n|x_{1:n})$.
The output action $\hat{x}_n$ is then sampled
from the distribution $p_\theta(x_n|z_n)$ of our conditional generative
model which is parameterized by $\theta$.
For mathematical convenience, we assume the action category and inter-arrival time are conditionally independent given the latent code $z_n$:
\begin{align}
p_\theta(x_n|z_n) = p_\theta(a_n, \tau_n |z_n) = p^a_\theta(a_n|z_n) p^\tau_\theta(\tau_n|z_n)
\end{align}
where $p^a_\theta(a_n|z_n)$ (resp. $p^\tau_\theta(\tau_n|z_n)$) is the conditional generative model for action category (resp. inter-arrival time).
This is a standard assumption in event prediction \cite{kdd_action_prediction, 2018arXiv180804063Z}.
The sequence model generates two probability distributions: (i) a categorical distribution over the action categories and (ii) a temporal point process distribution over the inter-arrival times for the next action. 

The distribution over action categories is modeled with a multinomial distribution when $a_n$ can only take a finite number of values:
\begin{equation}
p^a_\theta(a_n=k|z_n) = p_k(z_n) \quad \text{and} \,\,\,\,
\sum_{k=1}^K{p_k(z_n)} =1 \label{eq:action}
\end{equation}
where $p_k(z_n)$ is the probability of occurrence of action $k$, and $K$ is the total number of action categories. 

The inter-arrival time is assumed to follow an exponential distribution parameterized by $\lambda(z_n)$, similar to a standard temporal point process model:
\begin{equation}
\begin{aligned}
p^{\tau}_{\theta}(\tau_n | z_n) =
\begin{cases} 
\lambda(z_n) e^{-{\lambda(z_n)}\tau_n} & \text{if}~~ \tau_n \geq 0 \\
0 & \text{if}~~ \tau_n<0
\end{cases}
\end{aligned} \label{eq:time}
\end{equation}
where $p^{\tau}_{\theta}(\tau_n|z_n)$ is a probability density function over random variable $\tau_n$ and $\lambda(z_n)$ is the intensity of the process, which depends on the latent variable sample $z_n$.

\paragraph{Learning.}
We train the model by optimizing the variational lower bound over the entire sequence comprised of $N$ steps:
\begin{align}
\mathcal{L}_{\theta,\phi}(x_{1:N}) = \sum_{n=1}^N(&{\mathop{\mathbb{E}}}_{q_\phi(z_{n}|x_{1:n})}[\log p_\theta{(x_n|z_{n})}]  \\
&- D_{KL}(q_\phi(z_n|x_{1:n})||p_\psi(z_n|x_{1:n-1})))
\nonumber
\label{eq:loss}
\end{align}
Because the action category and inter-arrival time are conditionally independent given the latent code $z_n$, the log-likelihood term can be written as follows:
\begin{align}
&{\mathop{\mathbb{E}}}_{q_\phi(z_{n}|x_{1:n})}[\log p_\theta{(x_n|z_{n})}] = \\ &{\mathop{\mathbb{E}}}_{q_\phi(z_{n}|x_{1:n})}[\log p^a_\theta(a_n|z_{n})] 
+ {\mathop{\mathbb{E}}}_{q_\phi(z_{n}|x_{1:n})}[\log p^\tau_\theta(\tau_n|z_{n})]
\nonumber
\end{align}
Given the form of $p^a_\theta$ the log-likelihood term reduces to a cross entropy between the predicted action category distribution $p^a_\theta(a_n|z_{n})$ and the ground truth label $a^*_n$.
Given the ground truth inter-arrival time $\tau^*_n$, we compute its log-likelihood over a small time interval $\Delta_\tau$ under the predicted distribution.
\begin{align}
\log \left[\int_{\tau^*_n}^{\tau^*_n+\Delta_\tau}\!\!\!\!p^\tau_\theta(\tau_n|z_{n}) \!\!\dd\tau_n \right]
= \log(1 &- e^{-\lambda(z_{n}) \Delta_\tau}) \\&- \lambda(z_{n}) \tau^*_n \nonumber
\end{align}
We use the re-parameterization trick \cite{Kingma2014} to sample from the encoder network $q_\phi$.

\paragraph{Generation.} 
The goal is to generate the next action $\hat{x}_n=(\hat{a}_n, \hat{\tau}_n)$ given a sequence of past actions $x_{1:n-1}$.
The generation process is shown on the bottom of \autoref{fig:arch}.
At test time, an action at step $n$ is generated by first sampling
$z_n$ from the prior. 
The parameters of the prior distribution are computed based on the past $n-1$ actions $x_{1:n-1}$.
Then, an action category $\hat{a}_n$ and inter-arrival time $\hat{\tau}_n$ are generated as follows:
\begin{align}
\hat{a}_n \sim p_\theta^a(a_n|z_n) \qquad\qquad
\hat{\tau}_n \sim p_\theta^\tau(\tau_n|z_n)
\end{align}

\paragraph{Architecture.}
We now describe the architecture of our model in detail.
At step $n$, the current action $x_n$ is embedded into a vector representation $x_n^{emb}$ with a two-step embedding strategy.
First, we compute a representation for the action category ($a_n$) and the inter-arrival time ($\tau_n$) separately. 
Then, we concatenate these two representations and compute a new representation $x_n^{emb}$ of the action.
\begin{align}
a_n^{emb} = &f_a^{emb}(a_n) 
\qquad \qquad \tau_n^{emb} = f_\tau^{emb}(\tau_n)
\\
&x_n^{emb} = f_{a,\tau}^{emb}([a_n^{emb},\tau_n^{emb}])
\end{align}
We use a 1-hot encoding to represent the action category label $a_n$.
Then, we have two branches: one to estimate the parameters of the posterior distribution and another to estimate the parameters of the prior distribution. 
The network architecture of these two branches is similar but we use separate networks because the prior and the posterior distribution capture different information.
Each branch has a Long Short Term Memory (LSTM) \cite{Hochreiter1997} to encode the current action and the past actions into a vector representation:
\begin{align}
h_n^{post} &= LSTM_\phi(x_n^{emb}, h_{n-1}^{post}) \\
h_n^{prior} &= LSTM_\psi(x_n^{emb}, h_{n-1}^{prior}) 
\end{align}
Recurrent networks turn variable length sequences into meaningful, fixed-sized representations.
The output of the posterior LSTM $h_n^{post}$ (resp. prior LSTM $h_n^{prior}$) is passed into a posterior (also called inference) network $f^{post}_\phi$ (resp. prior network $f^{prior}_\psi$) that outputs the parameters of the Gaussian distribution:
\begin{align}
\mu_{\phi_n}, \sigma^2_{\phi_n} &= f^{post}_\phi(h_n^{post}) \\
\mu_{\psi_n}, \sigma^2_{\psi_n} &= f^{prior}_\psi(h_n^{prior}) 
\end{align}
Then, a latent variable ${z}_n$ is sampled from the posterior (or prior during testing) distribution and is fed to the decoder networks for generating distributions over the action category $a_n$ and inter-arrival time $\tau_n$.

The decoder network for action category $f^{a}_{\theta}(z_n)$ is a multi-layer perceptron with a softmax output to generate the probability distribution in Eq.~\ref{eq:action}:
\begin{equation}
p_\theta^a(a_n|z_n) = f^{a}_\theta({z}_n)
\end{equation}
The decoder network for inter-arrival time $f^{\tau}_{\theta}(z_n)$ is another multi-layer perceptron, producing the parameter for the point process model for temporal distribution in Eq.~\ref{eq:time}:
\begin{equation}
\lambda(z_n)= f^{\tau}_\theta({z}_n)
\end{equation}
During training, the parameters of all the networks are jointly learned in an end-to-end fashion.

\begin{table*}
\small
\centering
\begin{tabular}{llccccc}
\toprule
Dataset &Model  & Stoch. Var.  & LL  \\
\midrule
\multirow{3}{*}{Breakfast}&APP-LSTM   & -& -6.668  \\
&APP-VAE w/o Learned Prior & \cmark &$\geq$-9.427\\
&APP-VAE  & \cmark &$\geq$\textbf{-5.944}\\
\midrule
\multirow{3}{*}{MultiTUHMOS  }&APP-LSTM  & - &  -4.190  \\
&APP-VAE w/o Learned Prior & \cmark &$\geq$-5.344\\
&APP-VAE  & \cmark &$\geq$\textbf{-3.838}\\
\bottomrule
\end{tabular}
\vspace{0.05in}
\caption{Comparison of log-likelihood on Breakfast and MultiTHUMOS datasets.}
\label{tab:LogLikelihood}
\end{table*}
\section{Experiments}

\paragraph{Datasets.}
We performed experiments using APP-VAE on two action recognition datasets.  We use the standard training and testing sets for each.

\noindent \textit{MultiTHUMOS Dataset} \cite{Yeung2015} is a challenging dataset for action recognition, containing 400 videos of 65 different actions. On average, there are 10.5 action class labels per video and 1.5 actions per frame. 

\noindent \textit{Breakfast Dataset} \cite{Kuehne2014} contains 1712 videos of breakfast preparation for 48 action classes. The actions are performed by 52  people in 18 different kitchens.

\paragraph{Architecture details.} 
The APP-VAE model architecture is shown in Fig.\ \ref{fig:arch}.
Action category and inter-arrival time inputs are each passed through 2 layer MLPs with ReLU activation. They are then concatenated and followed with a linear layer. Hidden state of prior and posterior LSTMs is 128. Both prior and posterior networks are 2 layer MLPs, with ReLU activation after the first layer. Dimension of the latent code is 256. Action decoder is a 3 layer MLP with ReLU at the first two layers and softmax for the last one. The time decoder is also a 3 layer MLP with ReLU at the first two layers, with an exponential non-linearity applied to the output to ensure the parameter of the point process is positive.

\paragraph{Implementation details.}
The models are implemented with PyTorch \cite{Paszke2017} and are trained using the Adam \cite{kingma2014adam} optimizer for 1,500 epochs with batch size 32 and learning rate 0.001. 
We split the standard training set of both datasets into training and validation sets containing 70\% and 30\% of samples respectively. 
We select the best model during training based on the model loss (Eq.~\ref{eq:loss}) on the validation set.

\paragraph{Baselines.}
We compare APP-VAE with the following models for action prediction tasks. 
\begin{itemize}
\item \textit{Time Deterministic LSTM (TD-LSTM)}. 
This is a vanilla LSTM model that is trained to predict the next action category and the inter-arrival time, comparable with the model proposed by Farha~\etal~\cite{Farha2018}. This model directly predicts the inter-arrival time and not the distribution over it. TD-LSTM uses the same encoder network as APP-VAE.
We use cross-entropy loss for action category output and perform regression over inter-arrival time using mean squared error (MSE) loss similar to \cite{Farha2018}.
\item \textit{Action Point Process LSTM (APP-LSTM)}. 
This baseline predicts the inter-arrival time distribution similar to APP-VAE. The model uses the same reconstruction loss function as in the VAE model -- cross entropy loss for action category and negative log-likelihood (NLL) loss for inter-arrival time. APP-LSTM does not have the stochastic latent code that allows APP-VAE to model diverse distributions over action category and inter-arrival time. Our APP-LSTM baseline encompasses Du~\etal~\cite{kdd_action_prediction}'s work. The only difference is the way we model the intensity function (IF). Du~\etal~\cite{kdd_action_prediction} defines IS explicitly as a function of time. This design choice has been investigated in Zhong~\etal~\cite{2018arXiv180804063Z}; an implicit intensity function is shown to be superior and thus adapted in our APP-LSTM baseline. 
\end{itemize}

\paragraph{Metrics.} 
We use log-likelihood (LL) to compare our model with the APP-LSTM. 
We also report accuracy of action category prediction and mean absolute error (MAE) of inter-arrival time prediction. We calculate accuracy by comparing the most probable action category from the model output with the ground truth category. 
To calculate MAE, we use the expected inter-arrival time under the predicted distribution $p_{\theta}^{\tau}(\tau_n|z_n)$:
\begin{align}
\mathbb{E}_{p_{\theta}^{\tau}(\tau_n|z_n)}[\tau_n] &= \int\limits_0^\infty \tau_n\cdot p_{\theta}^{\tau}(\tau_n|z_n) \mathrm{d}\tau_n = \frac{1}{\lambda(z_n)}
\label{eq:expected_tau}
\end{align}
The expected value $\frac{1}{\lambda(z_n)}$ and the ground truth inter-arrival time are used to compute MAE.

\begin{figure*}[th]
\centering
\includegraphics[scale=.4]{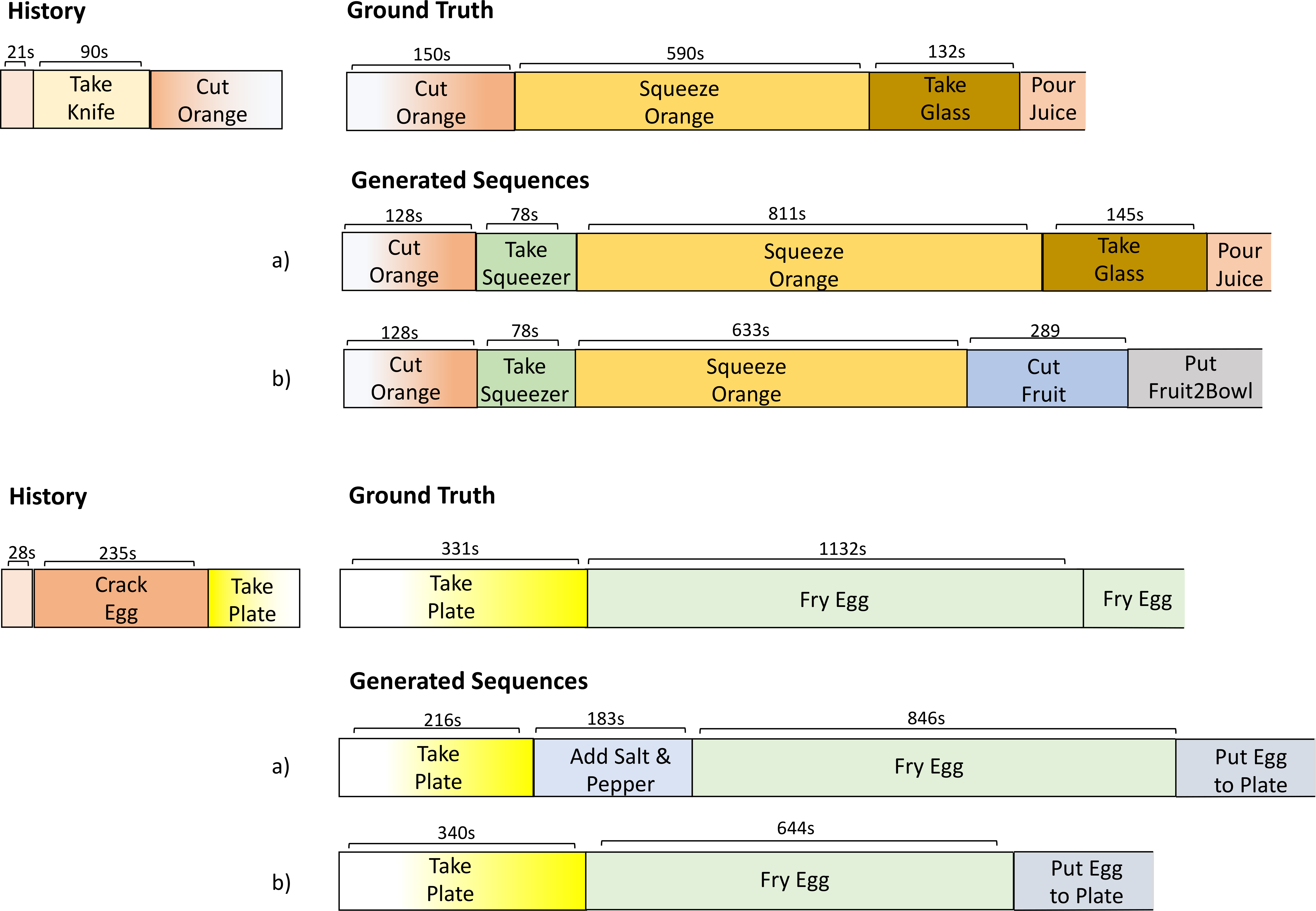}
\caption{Examples of generated sequences. Given the history (shown at left), we generate a distribution over latent code $z_n$ for the subsequent time step. A sample is drawn from this distribution, and decoded into distributions over action category and time, from which a next action/time pair by selecting the action with the highest probability and computing the expectation of the generated distribution over $\tau$ (\autoref{eq:expected_tau}). This process is repeated to generate a sequence of actions.  Two such sampled sequences (a) and (b) are shown for each history, and compared to the respective ground truth sequence (in line with history row). We can see that APP-VAE is capable of generating diverse and plausible action sequences.}
\label{fig:gen}
\end{figure*}

\begin{table*}[ht]
\centering
\begin{tabular}{llcccccccc}
\toprule
Dataset &Model & Time Loss & stoch. var.  & $\uparrow$ accuracy & $\downarrow$ MAE \\
\midrule
\multirow{4}{*}{Breakfast}&TD-LSTM  & MSE & - & 53.64 & 173.76 \\
&APP-LSTM  & NLL & - & 61.39 & 152.17 \\
&APP-VAE w/o Learned Prior & NLL & \cmark &27.09 & 270.75\\
&APP-VAE  & NLL & \cmark & \textbf{62.20} & \textbf{142.65} \\
\midrule
\multirow{4}{*}{MultiTUHMOS}&TD-LSTM & MSE & - & 29.74 & 2.33 \\
&APP-LSTM  & NLL & - & 36.31 & 1.99 \\
&APP-VAE w/o Learned Prior  & NLL & \cmark &8.79 & 2.02\\
&APP-VAE   & NLL & \cmark & \textbf{39.30} & \textbf{1.89} \\
\bottomrule
\end{tabular}
\caption{Accuracy of action category prediction and Mean Absolute Error (MAE) of inter-arrival time prediction of all model variants. Arrows show whether lower ($\downarrow$ ) or higher ($\uparrow$ ) scores are better.}
\label{tab:accuracy}
\end{table*}

\begin{table*}[ht]
\small
\centering
\newcolumntype{L}[1]{>{\hsize=#1\hsize\raggedright\arraybackslash}X}%
\newcolumntype{R}[1]{>{\hsize=#1\hsize\raggedleft\arraybackslash}X}%
\begin{tabularx}{0.95\linewidth}{L{0.001}L{2}}
\toprule
 & \textbf{Test sequences with high likelihood} \\
\midrule
1&   NoHuman,    CliffDiving,    Diving,    Jump,    BodyRoll,    CliffDiving,    Diving,    Jump,    BodyRoll,    CliffDiving,    Diving,    Jump,    BodyRoll,   BodyContract,    Run,    CliffDiving,    Diving,    Jump,  ..., BodyRoll, CliffDiving,    Diving,    BodyContract,    CliffDiving,     Diving,   CliffDiving,    Diving,    CliffDiving,    Diving,    Jump,    CliffDiving,    Diving,    Walk,    Run,    Jump,    Jump,    Run,    Jump\\
\midrule
2& CleanAndJerk,  PickUp,  BodyContract,  Squat,  StandUp,  BodyContract,  Squat,  CleanAndJerk,  PickUp,  StandUp,  BodyContract,   Squat,  CleanAndJerk,  PickUp,  StandUp,  Drop,  BodyContract, Squat,  PickUp, ..., Squat,  StandUp,  Drop,  BodyContract,  Squat, BodyContract,  Squat,  BodyContract,  Squat,  BodyContract,  Squat,  BodyContract,  Squat,  NoHuman \\
\midrule
& \textbf{Test sequences with low likelihood} \\
\midrule
1&  NoHuman,  TalkToCamera,  GolfSwing,  GolfSwing,  GolfSwing,  GolfSwing,  NoHuman \\
\midrule
2 & NoHuman, HammerThrow, TalkToCamera, CloseUpTalkToCamera, HammerThrow, HammerThrow, HammerThrow, TalkToCamera, ..., HammerThrow, HammerThrow, HammerThrow, HammerThrow, HammerThrow, HammerThrow, HammerThrow, HammerThrow, HammerThrow, HammerThrow, HammerThrow, HammerThrow, HammerThrow\\
\bottomrule
\end{tabularx}
\caption{Example of test sequences with high and low likelihood according to our learned model
}
\label{tab:highlow}
\end{table*}

\begin{figure}[t]
\includegraphics[width=1\linewidth]{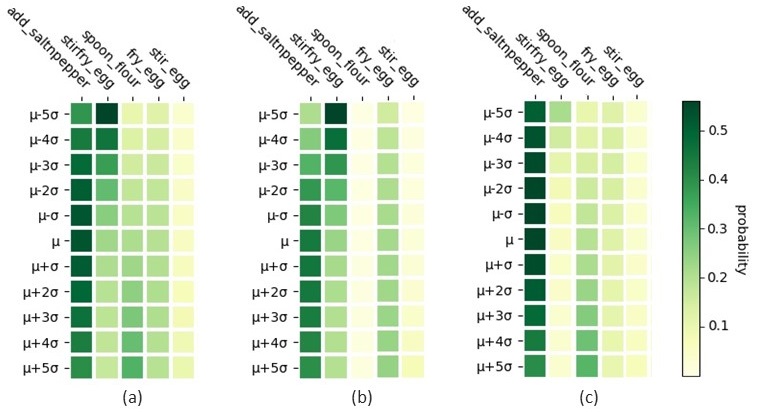} \\
\caption{\small{\textbf{Latent Code Manipulation.} The history + ground-truth label of future action for the sub-figures are: (a) ``SIL, crack\_egg"$\rightarrow$``add\_saltnpepper", (b) ``SIL, take\_plate, crack\_egg"$\rightarrow$ ``add\_saltnpepper" and (c) ``SIL, pour\_oil, crack\_egg"$\rightarrow$``add\_saltnpepper". } }
\label{fig:traversal}
\end{figure}

\begin{table}[ht]
\centering
\begin{tabular}{llccccccccc}
\toprule
Dataset & Model   & Acc & MAE \\
\midrule
Breakfast & APP-VAE - avg & 59.02 & 145.95\\
& APP-VAE - mode  & 62.20 & 142.65\\
\midrule
MultiTUHMOS & APP-VAE - avg  &35.23& 1.96\\
& APP-VAE - mode& 39.30 & 1.89 \\
\bottomrule
\end{tabular}
\caption{Accuracy (Acc) and Mean Absolute Error (MAE) under mode and averaging over samples.}
\label{tab:mode_avg}
\end{table}

\subsection{Experiment Results}

We discuss quantitative and qualitative results from our experiments. All quantitative experiments are performed by teacher forcing methodology \ie for each step in the sequence of actions, the models are fed the ground truth history of actions, and likelihood and/or other metrics for the next action are measured. 

\paragraph{Quantitative results.}
Table \ref{tab:LogLikelihood} shows experimental results that compare APP-VAE with the APP-LSTM. 
To estimate the log-likelihood (LL) of our model, we draw 1500 samples from the approximate posterior distribution, following the standard approach of importance sampling.
APP-VAE outperforms the APP-LSTM on both MultiTHUMOS and Breakfast datasets. We believe that this is because the APP-VAE model is better in modeling the complex distribution over future actions.

Table \ref{tab:accuracy} shows accuracy and MAE in predicting the future action given the history of previous actions. APP-VAE outperforms TD-LSTM and APP-LSTM under both the metrics. For each step in the sequence we draw 1500 samples from the prior distribution that models the next step action. 
Given the output distributions, we select the action category with the maximum probability as the predicted action, and the expected value of inter-arrival time as the predicted inter-arrival time. 
Out of 1500 predictions, we select the most frequent action as the model prediction for that time step, and compute inter-arrival time by averaging over the corresponding time values. 

Table \ref{tab:LogLikelihood} and \ref{tab:accuracy} also show the comparison of our model with the case where the prior is fixed in all of the time-steps. In this experiment, we fixed the prior to the standard normal distribution $\mathcal{N}(0,I)$. We can see that the learned prior variant outperforms the fixed prior variant consistently across all datasets. The model with the fixed prior does not perform well because it learns to predict the majority action class and average inter-arrival time of the training set, ignoring the history of any input test sequence.

In addition to the above strategy of selecting the mode action at each step, we also report action category accuracy and MAE obtained by averaging over predictions of all 1500 samples. We summarize these results in Table \ref{tab:mode_avg}.  

We next explore the architecture of our model by varying the sizes of the latent variable. Table \ref{tab:latent_size} shows the  log-likelihood of our model for different sizes of the latent variable. We see that as we increase the size of the latent variable, we can model a more complex latent distribution which results in better performance.

\paragraph{Qualitative Results.}
Fig.~\ref{fig:gen} shows examples of diverse future action sequences that are generated by APP-VAE given the history.  For different provided histories, sampled sequences of actions are shown.  We note that the overall duration and sequence of actions on the Breakfast Dataset are reasonable.  Variations, e.g.\ taking the juice squeezer before using it, adding salt and pepper before cooking eggs, are plausible alternatives generated by our model.

Fig.~\ref{fig:traversal} visualizes a traversal on one of the latent codes for three different sequences by uniformly sampling one $z$ dimension over $\big[ \mu -5\sigma, \mu+5\sigma \big]$ while fixing others to their sampled values. As shown, this dimension correlates closely with the action \textit{add\_saltnpepper}, \textit{strifry\_egg} and \textit{fry\_egg}.

We further qualitatively examine the ability of the model to score the likelihood of individual test samples.  We sort the test action sequences according to the average per time-step likelihood estimated by drawing 1500 samples from the approximate posterior distribution following the importance sampling approach.  High scoring sequences should be those that our model deems as ``normal" while low scoring sequences those that are unusual.  
Tab.~\ref{tab:highlow} shows some example of sequences with low and high likelihood on the MultiTHUMOS dataset.  We note that a regular, structured sequence of actions  such as jump, body roll, cliff diving for a diving action or body contract, squat, clean and jerk for a weightlifting action  receives high likelihood. However, repeated hammer throws or golf swings with no set up actions receives a low likelihood.

\begin{table}[t]
\centering
\begin{tabular}{lccccc}
\toprule
 \small{Latent} size &  {32} & 64 & 128 & 256 &512\\
 \midrule
 LL ($\geq$) & -4.486& -3.947 &-3.940 &-3.838&-4.098\\
\bottomrule
\end{tabular}
\caption{Log-likelihood for APP-VAE with different latent variable dimensionality on MultiTHUMOS.}
\label{tab:latent_size}
\end{table}

Finally we compare asynchronous APP-LSTM with a synchronous variant (with constant frame rate) on Breakfast dataset.
The synchronous model predicts actions one step at a time and the sequence is post-processed to infer the duration of each action.
 The performance is significantly worse for both MAE time (152.17 vs 1459.99) and action prediction accuracy (61.39\% vs 28.24\%). A plausible explanation is that LSTMs cannot deal with very long-term dependencies.

\section{Conclusion}

We presented a novel probabilistic model for point process data -- a variational auto-encoder that captures uncertainty in action times and category labels.  As a generative model, it can produce action sequences by sampling from a prior distribution, the parameters of which are updated based on neural networks that control the distributions over the next action type and its temporal occurrence.  The model can also be used to analyze given input sequences of actions to determine the likelihood of observing particular sequences.  We demonstrate empirically that the model is effective for capturing the uncertainty inherent in tasks such as action prediction and anomaly detection.

\clearpage
{\small
\bibliographystyle{ieee_fullname}
\bibliography{egbib}
}

\clearpage
\appendix

\end{document}